\documentclass{article}

\usepackage{times}

\usepackage{hyperref}

\usepackage[accepted]{icml2015}

\usepackage{url}

\usepackage{amsfonts,amstext,amsmath,amssymb}

\usepackage{longtable}

\usepackage{graphicx}

\usepackage{array}

\usepackage{tikz}
\usetikzlibrary{bayesnet}

\usepackage{booktabs} 
\usepackage{multirow}
\newcommand{\ra}[1]{\renewcommand{\arraystretch}{#1}}

\icmltitlerunning{Phrase-based Image Captioning}

\begin{document}

\twocolumn[
\icmltitle{Phrase-based Image Captioning}

\icmlauthor{
R\'emi Lebret$^{1}$ \& Pedro O. Pinheiro\footnote{These two authors contributed equally to this work.}}{remi@lebret.ch, pedro@opinheiro.com}
\icmladdress{Idiap Research Institute, Martigny, Switzerland\\
\'Ecole Polytechnique F\'ed\'erale de Lausanne (EPFL), Lausanne, Switzerland}
\icmlauthor{Ronan Collobert$^{2}$}{ronan@collobert.com}
\icmladdress{Facebook AI Research, Menlo Park, CA, USA}
\icmlkeywords{Learning, Natural Language Processing, Computer Vision}

\vskip 0.3in 
]

\begin{abstract}
Generating a novel textual description of an image is an interesting problem that connects computer vision and natural language processing.  
In this paper, we present a simple model that is able to generate descriptive sentences given a sample image. 
This model has a strong focus on the syntax of the descriptions.
We train a purely bilinear model that learns a metric between an image representation (generated from a previously trained Convolutional Neural Network) and phrases that are used to described them. The system is then able to infer phrases from a given image sample. Based on caption syntax statistics, we propose a simple language model that can produce relevant descriptions for a given test image using the phrases inferred. Our approach, which is considerably simpler than state-of-the-art models, achieves comparable results in two popular datasets for the task: Flickr30k and the recently proposed Microsoft COCO.
\end{abstract}

\section{Introduction}
\label{intro}
Being able to automatically generate a description from an image is a fundamental problem in artificial intelligence, connecting computer vision and natural language processing. The problem is particularly challenging because it requires to correctly recognize different objects in images and how they interact. Another challenge is that an image description generator needs to express these interactions in a natural language (\emph{e.g.} English). Therefore, a language model is implicitly required in addition to visual understanding.

Recently, this problem has been studied by many different authors. Most of the attempts are based on recurrent neural networks to generate sentences. These models leverage the power of neural networks to transform image and sentence representations into a common space~\citep{MaoXYWY14,Andrej2014,VinyalsTBE14,DonahueHGRVSD14}. 

In this paper, we propose a different approach to the problem that does not rely on complex recurrent neural networks.
An exploratory analysis of two large datasets of image descriptions reveals that their syntax is quite simple.
The ground-truth descriptions can be represented as a collection of noun, verb and prepositional phrases. 
The different objects in a given image are described by the noun phrases, while the interactions between these objects are encoded by both the verb and the prepositional phrases. 
We thus train a model that predicts the set of phrases present in the sentences used to describe the images.
By leveraging previous works on word vector representations, each phrase can be represented by the mean of the representations of the words that compose the phrase. Vector representations for images can also be easily obtained from some pre-trained convolutional neural networks.
The model then learns a common embedding between phrase and image representations (see Figure~\ref{fig:schema}). 

Given a test image, a bilinear model is trained to predict a set of top-ranked phrases that best describe it. Several noun phrases, verb phrases and prepositional phrases are in this set. The objective is therefore to generate syntactically correct sentences from (possibly different) subsets of these phrases. 
We introduce a trigram constrained language model based on our knowledge about how the sentence descriptions are structured in the training set. 
With a very constrained decoding scheme, sentences are inferred with a beam search.
Because these sentences are not conditioned to the given image (apart with the initial phrases selection), a re-ranking is used to pick the sentence that is closest to the sample image (according to the learned metric).
The quality of our sentence generation is evaluated on two very popular datasets for the task: Flickr30k~\citep{HodoshYH13} and the recently published COCO~\citep{mscoco2014}. Using the popular BLEU score~\citep{Papineni:2002}, our results are competitive with other recent works. Our generated sentences also achieve a similar performance as humans on the BLEU metric.

The paper is organized as follows. Section~\ref{related-works} presents related works. Section~\ref{dataanalysis} presents the analysis we conducted to better understand the syntax of image descriptions. Section~\ref{model} describes the proposed phrase-based model. Section~\ref{sec:generation} introduces the sentence generation from the predicted phrases. Section~\ref{exp-results} describes our experimental setup and the results on the two datasets. Section~\ref{conclusion} concludes.

\section{Related Works}
\label{related-works}

The classical approach to sentence generation is to pose the problem as a retrieval problem: a given test image will be described with the highest ranked annotation in the training set \citep{HodoshYH13,SocherKLMN14,srivastava14b}. 
These matching methods may not generate proper descriptions for a new combination of objects.  
Due to this limitation, several generative approaches have been proposed. 
Many of them use syntactic and semantic constraints in the generation process~\citep{YaoYLLZ10,Mitchell:2012,Kulkarni:2011,Kuznetsova:2012}. These approaches benefit from visual recognition systems to infer words or phrases, but in contrast to our work they do not leverage a multimodal metric between images and phrases.

More recently, automatic image sentence description approaches based on deep neural networks have emerged with the release of new large datasets. As starting point, these solutions use the rich representation of images generated by Convolutional Neural Networks \citep{lecun1998gradient} (CNN) that were previously trained for object recognition tasks. 
These CNN are generally followed by recurrent neural networks (RNN) in order to generate full sentence descriptions~\cite{VinyalsTBE14,Andrej2014,DonahueHGRVSD14,Chen14,MaoXYWY14,VenugopalanXDRMS14,KirosSZ14}. Among these recent works, long short-term memory (LSTM) is often chosen as RNN.
In such approaches, the key point is to learn a common space between images and words or between images and sentences, i.e. a multimodal embedding.

\citet{VinyalsTBE14} consider the problem in a similar way as a machine translation problem. The authors propose an encoder/decoder (CNN/LSTM networks) system that is trained to maximize the likelihood of the target description sentence given a training image. \citet{Andrej2014} propose an approach that is a combination of CNN, bidirectional RNN over sentences and a structured objective responsible for a multimodal embedding. They then propose a second RNN architecture to generate new sentences.
Similarly, \citet{MaoXYWY14} and \citet{DonahueHGRVSD14} propose a system that uses a CNN to extract image features and a RNN for sentences. The two networks interact with each other in a multimodal common layer.

Our model shares some similarities with these recent proposed approaches. We also use a pre-trained CNN to extract image features. However, thanks to the phrase-based approach, our model does not rely on complex recurrent networks for sentence generation, and we do not fine-tune the image features.

As our approach, \citet{FangGISDDGHMPZZ14} proposes to not use recurrent networks for generating the sentences.
Their solution can be divided into three steps: (i) a visual detector for words that commonly occur are trained using multiple instance learning, (ii) a set of sentences are generated using a Maximum-Entropy language model and (iii) the set of sentences is re-ranked using sentence-level features and a proposed deep multimodal similarity model. Our work differs from this approach in two different important ways: our model infers phrases present in the sentences instead of words and we use a considerably simpler language model.


\section{Syntax Analysis of Image Descriptions}
\label{dataanalysis}

The art of writing sentences can vary a lot according to the domain. 
When reporting news or reviewing an item, not only the choice of the words might vary, but also the general structure of the sentence.
In this section, we wish to analyze the syntax of image descriptions to identify whether images have their own structures.
We therefore proceed to an exploratory analysis of two recent datasets containing a large amount of images with descriptions: Flickr30k \citep{HodoshYH13} and COCO \citep{mscoco2014}.

\subsection{Datasets} 
The Flickr30k dataset contains 31,014 images where 1,014 images are for validation, 1,000 for testing and the rest for training (i.e. 29,000 images). 
The COCO dataset contains 123,287 images, 82,783 training images and 40,504 validation images. The testing images has not yet been released.
We thus use two sets of 5,000 images from the validation images for validation and test, as in \citet{Andrej2014}\footnote{Available at \small\url{http://cs.stanford.edu/people/karpathy/deepimagesent/}}.
In both datasets, images are given with five (or six) sentence descriptions annotated using Amazon Mechanical Turk.  
This results in 559,113 sentences when combining both training datasets.


\subsection{Chunking-based Approach}

A quick overview over these sentence descriptions reveals that they all share a common structure, usually describing the different objects present in the image and how they interact between each other. 
This interaction among objects is described as actions or relative position between different objects. 
The sentence can be short or long, but it generally respects this process. 
To confirm this claim and better understand the description structures, we used a chunking (also called shallow parsing) approach which identifies the constituents of a sentence.
These constituents are usually noun phrases (NP), verb phrases (VP) and prepositional phrases (PP).
We extract them from the training sentences with the SENNA software\footnote{Available at \small\url{http://ml.nec-labs.com/senna/}}.
Pre-verbal and post-verbal adverb phrases are merged with verb phrases to limit the number of phrase types. 

\begin{figure}
  \includegraphics[width=\columnwidth]{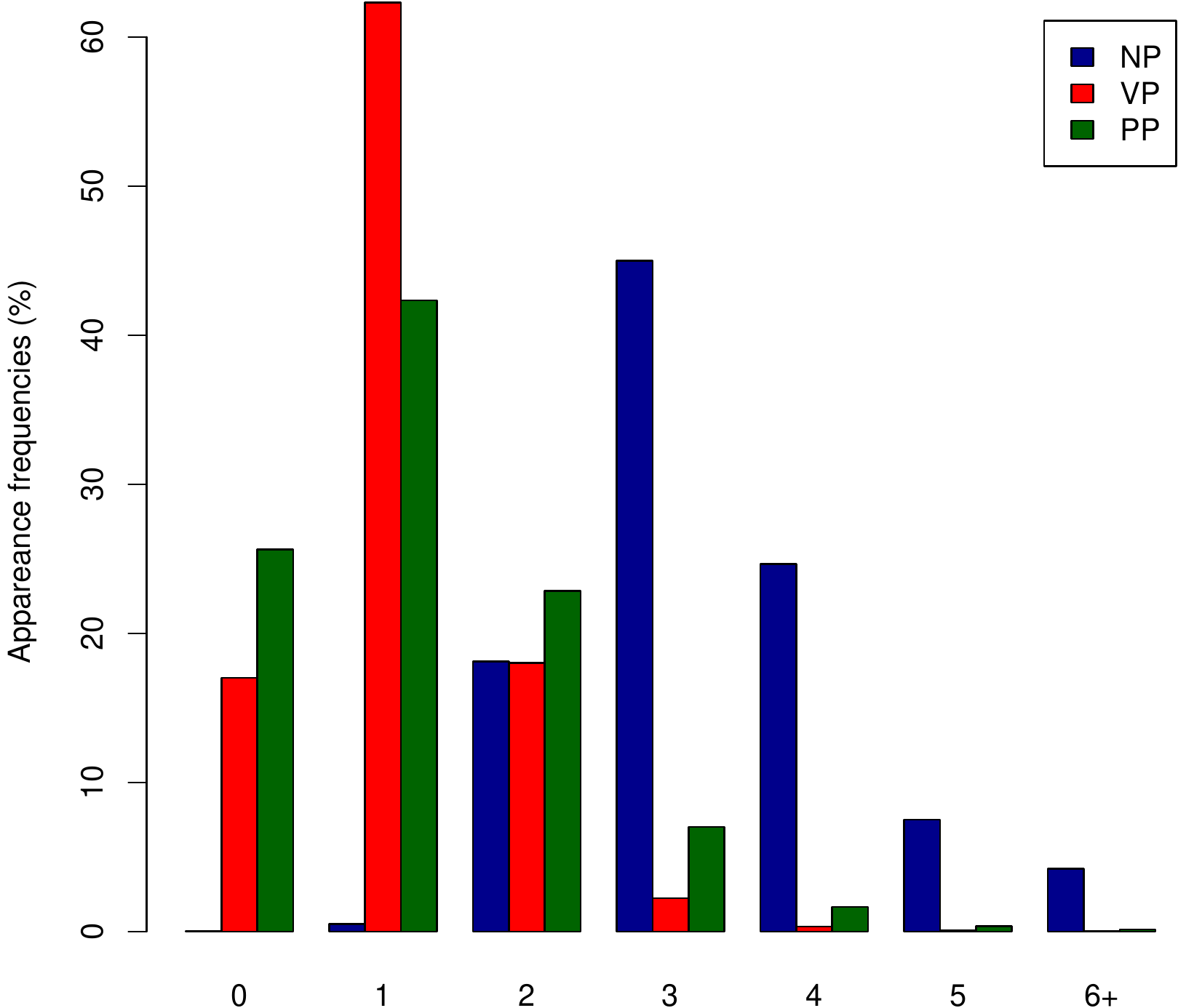}
  \caption{Statistics on the number of phrases (NP, VP, PP) per ground-truth descriptions in Flickr30k and COCO training datasets.}
  \label{fig:stats1}
\end{figure}

\begin{figure}[t]
  \includegraphics[width=\columnwidth]{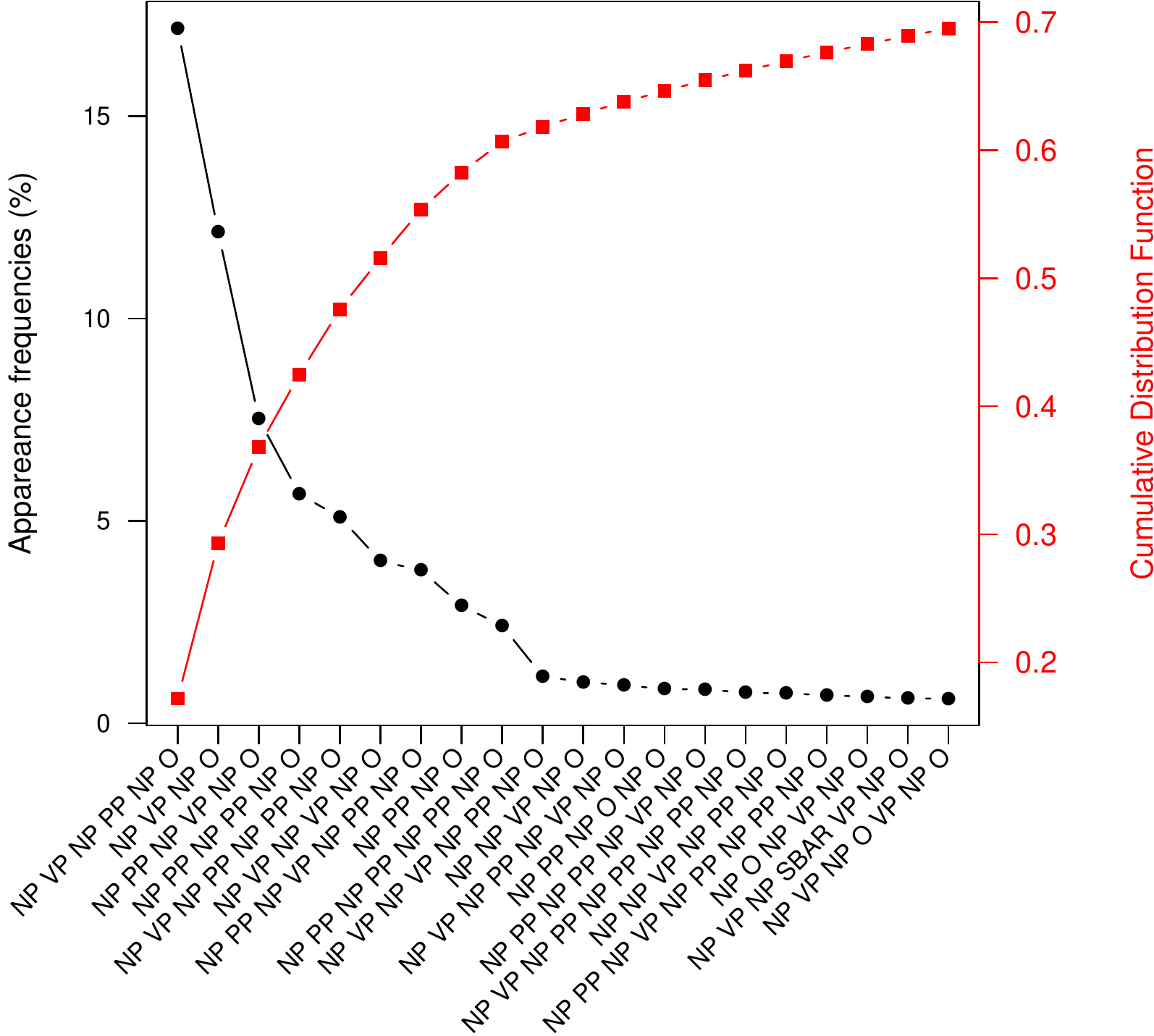}
  \caption{The 20 most frequent sentence structures in Flickr30k and COCO training datasets. The black line is the appearance frequency for each structure, the red line is the cumulative distribution.}
  \label{fig:stats2}
\end{figure}

Statistics reported in Figure~\ref{fig:stats1} and Figure~\ref{fig:stats2} confirm that image descriptions possess a simple and distinct structure. 
These sentences do not have much variability. 
All the key elements in a given image are usually described with a noun phrase (NP).
Interactions between these elements can then be explained using prepositional phrases (PP) or verb phrases (VP).
A large majority of sentences contain from two to four noun phrases. Two noun phrases then interact using a verb or prepositional phrase.
Describing an image is therefore just a matter of identifying these constituents.
We thus propose to train a model which can predict the phrases which are likely to be in a given image.

\section{Phrase-based Model for Image Descriptions}
\label{model}

\begin{figure*}[ht]
\centering
\includegraphics[height=7cm]{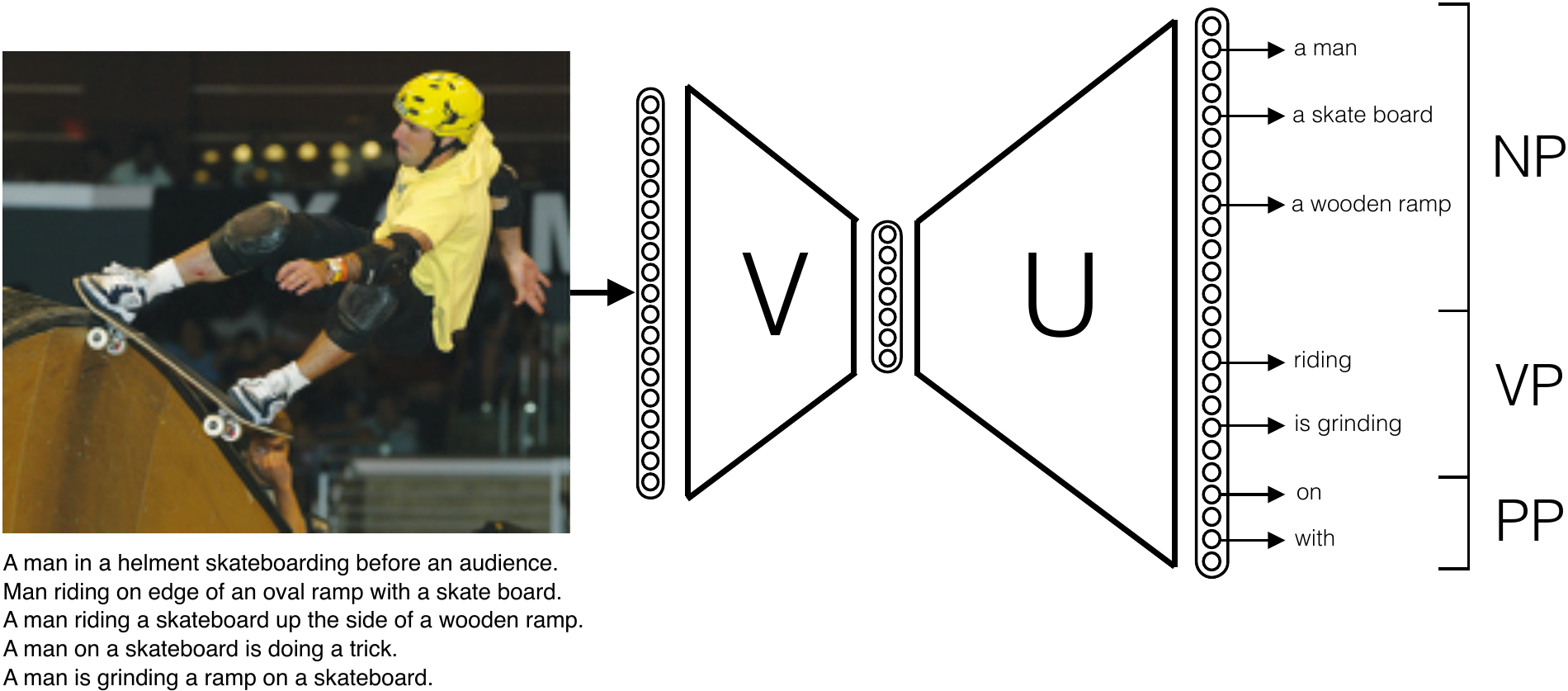}
\caption{Schematic illustration of our phrase-based model for image descriptions.}
\label{fig:schema}
\end{figure*}

By leveraging previous works on word and image representations, we propose a simple model which can predict the phrases that best describe a given image. For this purpose, a metric between images and phrases is trained, as illustrated in Figure~\ref{fig:schema}. The proposed architecture is then just a low-rank bilinear model $U^TV$.

\subsection{Image Representations}

For the representation of images, we choose to use a Convolutional Neural Network. CNN have been widely used in different vision domains and are currently the state-of-the-art in many object recognition tasks. We consider a CNN that has been pre-trained for the task of object classification~\cite{Chatfield14}. We use a CNN solely to the purpose of feature extraction, that is, no learning is done in the CNN layers.

\subsection{Learning a Common Space for Image and Phrase Representations}

Let $\mathcal{I}$ be the set of training images, $\mathcal{C}$ the set of all phrases used to describe $\mathcal{I}$, and $\theta$ the trainable parameters of the model.
By representing each image $i \in \mathcal{I}$ with a vector $\mathrm{z}_i \in \mathbb{R}^n$ thanks to the pre-trained CNN, we define a metric between the image $i$ and a phrase $c$ as a bilinear operation:
\begin{equation}\label{eq:score}
f_\theta(c,i) =  \mathrm{u}_c^T V \mathrm{z}_i \,,
\end{equation}
with $U= (\mathrm{u}_{c_1}, \ldots, \mathrm{u}_{c_{|\mathcal{C}|}}) \in \mathbb{R}^{m \times |\mathcal{C}|}$ and $V\in \mathbb{R}^{ m \times n}$ being the trainable parameters $\theta$.
Note that $U^TV$ could be a full matrix, but a low-rank setting eases the capacity control.

\subsection{Phrase Representations Initialization}

Noun phrases or verb phrases are often a combination of several words.
Good word vector representations can be obtained very efficiently with many different recent approaches~\citep{Mikolov2013,Mnih2013,pennington2014glove,Lebret14b}.
\citet{MikolovICLR2013} also showed that simple vector addition can often produce meaningful results, such as \emph{king - man + woman $\approx$ queen}.
By leveraging the ability of these word vector representations to compose by simple summation, representations for phrases are easily computed with an element-wise addition.

Each phrase $c$ composed of $K$ words $w_k$ is therefore represented by a vector $\mathrm{x}_{w_k} \in \mathbb{R}^m$ thanks to a word representation model pre-trained on large unlabeled text corpora.
A vector representation $\mathrm{u}_c$ for a phrase $c = \{w_1,\ldots,w_K\}$ is then calculated by averaging its word vector representations:
\begin{equation}
 \mathrm{u}_{c} = \frac{1}{K} \sum_{k=1}^K \mathrm{x}_{w_k}\,.
\end{equation}
Vector representations for all phrases $c \in \mathcal{C}$ can thus be obtained to initialized the matrix $U \in \mathbb{R}^{m \times |\mathcal{C}|}$.
$V \in \mathbb{R}^{ m \times n}$ is initialized randomly and trained to encode images in the same vector space than the phrases used for their descriptions. 

\subsection{Training with Negative Sampling}
Each image $i$ is described by a multitude of possible phrases $\mathcal{C}^{i}\subseteq \mathcal{C}$.
We consider $|\mathcal{C}|$ classifiers attributing a score for each phrase.
We train our model to discriminate a target phrase $c_j$ from a set of negative phrases $c_k \in \mathcal{C}^{-} \subseteq \mathcal{C}$, with $c_k \neq c_j$.
With $\theta=\{U,V\}$, we minimize the following logistic loss function with respect to $\theta$:
\begin{multline}
\theta \mapsto  \sum_{i \in \mathcal{I}} \sum_{c_{j} \in \mathcal{C}^{i}}  \Big( \log \Big(1 +e^{-\mathrm{u}_{c_j}^T V \mathrm{z}_i} \Big) \\+ \sum_{c_k \in \mathcal{C}^{-}} \log \Big(1 +e^{+\mathrm{u}_{c_k}^T V \mathrm{z}_i}\Big)  \Big)\,.
\end{multline}
The model is trained using stochastic gradient descent. A new set of negative phrases $\mathcal{C}^{-}$ is randomly picked from the training set at each iteration.

\section{From Phrases to Sentence}
\label{sec:generation}

After identifying the $L$ most likely constituents $c_j$ in the image $i$, we propose to generate sentences out of them.
From this set, $l \in \{1,\ldots,L\}$ phrases are used to compose a syntactically correct description. 

\subsection{Sentence Generation}
\label{sec:markov}
Using a statistical language framework, the likelihood of a certain sentence is given by:
\begin{equation}\label{eq:ngram}
P(c_1,c_2,\ldots,c_l) = \prod_{j=1}^l P(c_j | c_1,\ldots,c_{j-1})
\end{equation}
Keeping this system as simple as possible and using the second order Markov property, we approximate Equation~\ref{eq:ngram} with a trigram language model: 
\begin{equation}\label{eq:trigram}
P(c_1,c_2,\ldots,c_l) \approx \prod_{j=1}^l P(c_j | c_{j-2}, c_{j-1})\,.
\end{equation}
The best candidate corresponds to the sentence $P(c_1,c_2,\ldots,c_l)$ which maximizes the likelihood of Equation~\ref{eq:trigram} over all the possible sizes of sentence.
Because we want to constrain the decoding algorithm to include prior knowledge on chunking tags $t \in \{NP,VP,PP\}$, we rewrite Equation~\ref{eq:trigram} as:
\begin{align}
&\prod_{j=1}^l  \sum_{t} P(c_j | t_j = t, c_{j-2}, c_{j-1})P(t_j = t | c_{j-2}, c_{j-1})\nonumber\\
&= \prod_{j=1}^l P(c_j | t_j, c_{j-2}, c_{j-1})P(t_j| c_{j-2}, c_{j-1})\,.
\end{align}
Both conditions $P(c_j | t_j, c_{j-2}, c_{j-1})$ and $P(t_j| c_{j-2}, c_{j-1})$ are probabilities estimated by counting trigrams in the training datasets.

\subsection{Sentence Decoding}

\begin{figure}[h!]
  \begin{center}
%
%
%
%

\begin{tikzpicture}

  \node[obs]                  (np) {NP};
  \node[latent, right=1cm of np, yshift=0cm] (c1) {$c$};
  \node[obs, right=1cm of c1, yshift=1.5cm] (vp) {VP};
  \node[obs, right=1cm of c1, yshift=0cm] (pp) {PP};
  \node[obs, right=1cm of c1, yshift=-1.5cm] (period) {.};
  \node[latent, right=1 cm of pp, yshift=0.75cm] (c2)  {$c$} ; %

  \node[const, left=0.5 cm of np, yshift=-0.0cm] (start)  {start} ; %
  \node[const, above=1.25 cm of c2] (tmp2)  {} ; %
  \node[const, left= 5.1 cm of tmp2] (tmp3)  {} ; %

  \edge {start} {np} ; %
  \edge {np} {c1} ; %
  \edge {c1} {vp,pp,period} ; %
  \edge {vp,pp} {c2} ; %
  \myline {c2} {tmp2} ; %
  \myline {tmp2} {tmp3} ; %
  \edge {tmp3} {np} ; %

  \plate {} {(np)(vp)(pp)(period)(c1)(c2)} {$N$} ;

\end{tikzpicture}
  \end{center}
  \caption{The constrained language model for generating description given the predicted phrases for an image.}
  \label{fig:bayes}
\end{figure}
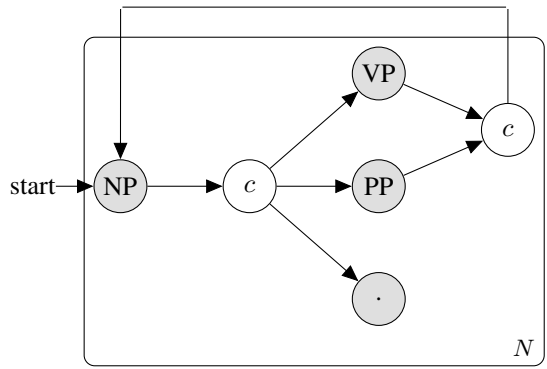

At decoding time, we prune the graph of all possible sentences made out of the top $L$ phrases with a beam search, according to three heuristics:
(i) we consider only the transitions which are likely to happen (we discard any sentence which would have a trigram transition probability inferior to 0.01). This thresholding helps to discard sentences that are semantically incorrect;
(ii) each predicted phrases $c_j$ may appear only once\footnote{This is easy to implement with a beam search, but intractable with a full search.};
(iii) we add syntactic constraints which are illustrated in Figure~\ref{fig:bayes}.
The last heuristic is based on the analysis of syntax in Section~\ref{dataanalysis}.
In Figure~\ref{fig:stats2}, we see that a noun phrase is, in general, always followed by a verb phrase or a prepositional phrase, and both are then followed by another noun phrase. A large majority of the sentences contain three noun phrases interleaved with verb phrases or prepositional phrases.
According the statistics reported in Figure~\ref{fig:stats1}, sentences with two or four noun phrases are also common, but sentences with more than four noun phrases are marginal.
We thus repeat this process $N=\{2,3,4\}$ times until reaching the end of a sentence (characterized by a period).

\subsection{Sentence Re-ranking}
For each test image $i$, the proposed model will generate a set of $M$ sentences.
Sentence generation is not conditioned on the image, apart from phrases which are selected beforehand.
Some phrase sequences might be syntactically good, but have low match with the image.
Consider, for instance, an image with a cat and a dog. Both sentences \emph{``a cat sitting on a mat and a dog eating a bone''} and \emph{``a cat sitting on a mat''} are correct, but the second is missing an important part of the image.
A ranking of the generated sentences is therefore necessary to choose the one that has the best match with the image. 

Because a generated sentence is composed from $l$ phrases predicted by our system, we simply average the phrase scores given by Equation~\ref{eq:score}. 
For a generated sentence $s$ composed of $l$ phrases $c_j$, a score between $s$ and $i$ is calculated as: 
\begin{equation}
\frac{1}{l} \sum_{c_j \in s} f_{\theta}(c_j,i)\,.
\end{equation}
The best candidate is the sentence which has the highest score out of the $M$ generated sentences. 
This ranking helps the system to chose the sentence which is closer to the sample image.

\section{Experiments}

\label{exp-results}
\subsection{Experimental Setup}


\subsubsection{Feature Selection} 
Following \citet{Andrej2014}, the image features are extracted using VGG CNN \citep{Chatfield14}. This model generates image representations of dimension 4096 from RGB input images. 

\begin{table}[h!]
\ra{1.3}
\begin{center}
\begin{tabular}{@{}lcccc@{}}
\hline\toprule
& \phantom{a} & {\bf Flickr30k} & \phantom{a} & {\bf COCO} \\
\bottomrule
Noun Phrase (NP) & & 4818 & & 8982 \\
Verb Phrase (VP) & & 2109 & & 3083\\
Prepositional Phrase (PP) & & 128 & & 189\\
\midrule
Total $\mathcal{|C|}$ & & 7055 & & 12254 \\ 
\bottomrule
\hline
\end{tabular}
\end{center}
\caption{Statistics of phrases appearing at least ten times.}
\label{tab:stats}
\end{table}
For each training set, only phrases occurring at least ten times are considered.
This threshold is chosen to fulfil two objectives: (i) limit the number of phrases $\mathcal{C}$ and therefore the size of the matrix $U$ and (ii) exclude rare phrases to better generalize the descriptions. Statistics on the number of phrases are reported in Table~\ref{tab:stats}. 
For Flickr30k, this threshold covers about 81\% of NP, 83\% of VP and 99\% of PP. For COCO, it covers about 73\% of NP, 75\% of VP and 99\% of PP.
Phrase representations are then computed by averaging vector representations of their words.
We obtained word representations from the Hellinger PCA of a word co-occurrence matrix, following the method described in~\citet{Lebret14b}.
The word co-occurrence matrix is built over the entire English Wikipedia\footnote{Available at \small\url{http://download.wikimedia.org}. We took the January 2014 version.}, with a symmetric context window of ten words coming from the 10,000 most frequent words. Words, and therefore also phrases, are represented in 400-dimensional vectors.

\subsubsection{Learning the Multimodal Metric}
The parameters $\theta$ are $V \in \mathbb{R}^{400 \times 4096 }$ (initialized randomly) and $U \in \mathbb{R}^{400 \times |\mathcal{C}|}$ (initialized with the phrase representations) which are tuned on the validation datasets. They are trained with $15$ randomly chosen negative samples  and a learning rate set to 0.00025.

\subsubsection{Generating Sentences from the Predicted Phrases}
\label{sec:generatingSentences}
Transition probabilities for our constrained language model (see Figure~\ref{fig:bayes}) are calculated independently for each training set.
No smoothing has been used in the experiments.
Concerning the set of top-ranked phrases for a given test image, we select only the top five predicted verb phrases and the top five predicted prepositional phrases.
Since the average number of noun phrases is higher than for the two other types of phrases (see Figure~\ref{fig:stats1}), more noun phrases are needed. The top twenty predicted noun phrases are thus selected. 

\subsection{Experimental Results} 

As a first evaluation, we consider the task of retrieving the ground-truth phrases from test image descriptions.
Results reported in Table~\ref{tab:recall} show that our system achieves a recall of around 50\% on this task on the test set of both datasets, assuming the threshold considered for each type of phrase (see \ref{sec:generatingSentences}). Note that this task is extremely difficult, as semantically similar phrases (\emph{the women} / \emph{women} / \emph{the little girls}) are classified separately.
Despite the possible number of noun phrases being higher, results in Table~\ref{tab:recall} reveal that noun phrases are better retrieved than verb phrases.
This shows that our system is able to detect different objects in the image. However, finding the right verb phrase seems to be more difficult. 
A possible explanation could be that there exists a wide choice of verb phrases to describe interactions between the noun phrases.
For instance, we see in Figure~\ref{fig:schema} that two annotators have used the same noun phrases (\emph{a man}, \emph{a skateboard} and \emph{a (wooden) ramp}) to describe the scene, but they have then chosen a different verb phrase to link them (\emph{riding} versus \emph{is grinding}).
Therefore, we suspect that a low recall for verb phrases does not necessarily mean that the predictions are wrong.
Finding the right prepositional phrase seems, on the contrary, much easier. 
The high recall for prepositional phrase can be explained by much lower variability of this type of phrase compared to the two others (see Table~\ref{tab:stats}). 

\begin{table}[h!]
\ra{1.3}
\begin{center}
\begin{tabular}{@{}lcccc@{}}
\hline\toprule
& \phantom{a} & {\bf Flickr30k} & \phantom{a} & {\bf COCO} \\
\bottomrule
Noun Phrase (NP) & & 38.14 & & 45.44 \\
Verb Phrase (VP) & & 20.61 & & 27.83 \\
Prepostional Phrase (PP) & & 81.70 & & 84.49\\
\midrule
Total & & 44.92 & & 52.49 \\ 
\bottomrule
\hline
\end{tabular}
\end{center}
\caption{Recall on phrase retrieval. For each test image, we take the top 20 predicted NP, the top 5 predicted VP, and the top 5 predicted PP.}
\label{tab:recall}
\end{table}

\begin{table*}[ht!]
\ra{1.3}
\begin{center}
\begin{tabular}{@{}lccccccccccc@{}}
\hline\toprule
& \phantom{abc} & \multicolumn{4}{c}{\bf Flickr30K} & \phantom{abc} & \multicolumn{4}{c}{\bf COCO} \\
\cmidrule{3-6} \cmidrule{8-11}
& & B-1 & B-2 & B-3 & B-4 & & B-1 & B-2 & B-3 & B-4\\
\bottomrule
Human agreement  & & $0.55$ & $0.35$ & $0.23$ & $0.15$ & & $0.68$ & $0.45$ & $0.30$ & $0.20$\\
\midrule
\citet{MaoXYWY14} & & $0.55$ & $0.24$ & $0.19$ & - & & - & - & - & - \\
\citet{Andrej2014} &  & $0.50$ & $0.30$ & $0.15$ & - & & $0.57$ & $0.37$ & $0.19$ & -\\
\citet{VinyalsTBE14} & & $0.66$ & - & - & - & & $0.67$ & - & - & - \\
\citet{DonahueHGRVSD14} & & $0.59$ & $0.39$ & $0.25$ & $0.16$ & &  $0.63$ & $0.44$ & $0.30$ & $0.21$ \\
\citet{FangGISDDGHMPZZ14} & & - & - & - & - & & - & - & - & $0.21$ \\
Our model & & $0.59$ & $0.35$ & $0.20$ & $0.12$ & & $0.70$ & $0.46$ & $0.30$ & $0.20$ \\
\bottomrule
\hline
\end{tabular}
\end{center}
\caption{Comparison between human agreement scores, state of the art models and our model on both datasets. Note that there are slight variations between the test sets chosen in each paper.}
\label{tab:results}
\end{table*}


As a second evaluation, we consider the task of generating full descriptions. We measure the quality of the generated sentences using the popular, yet controversial, BLEU score~\citep{Papineni:2002}.
Table~\ref{tab:results} shows our sentence generation results on the two datasets considered. 
BLEU scores are reported up to 4-gram. Human agreement scores are computed by comparing the first ground-truth description against the four others\footnote{For all models, BLEU scores are computed against five reference sentences which give a slight advantage compared to human scores.}. 
For comparison, we include results from recently proposed models.
Our model, despite being simpler, achieves similar results to state of the art results. It is interesting to note that our results are very close to the human agreement scores.

\begin{figure*}[!t]
  \begin{center}
    \includegraphics[width=\textwidth]{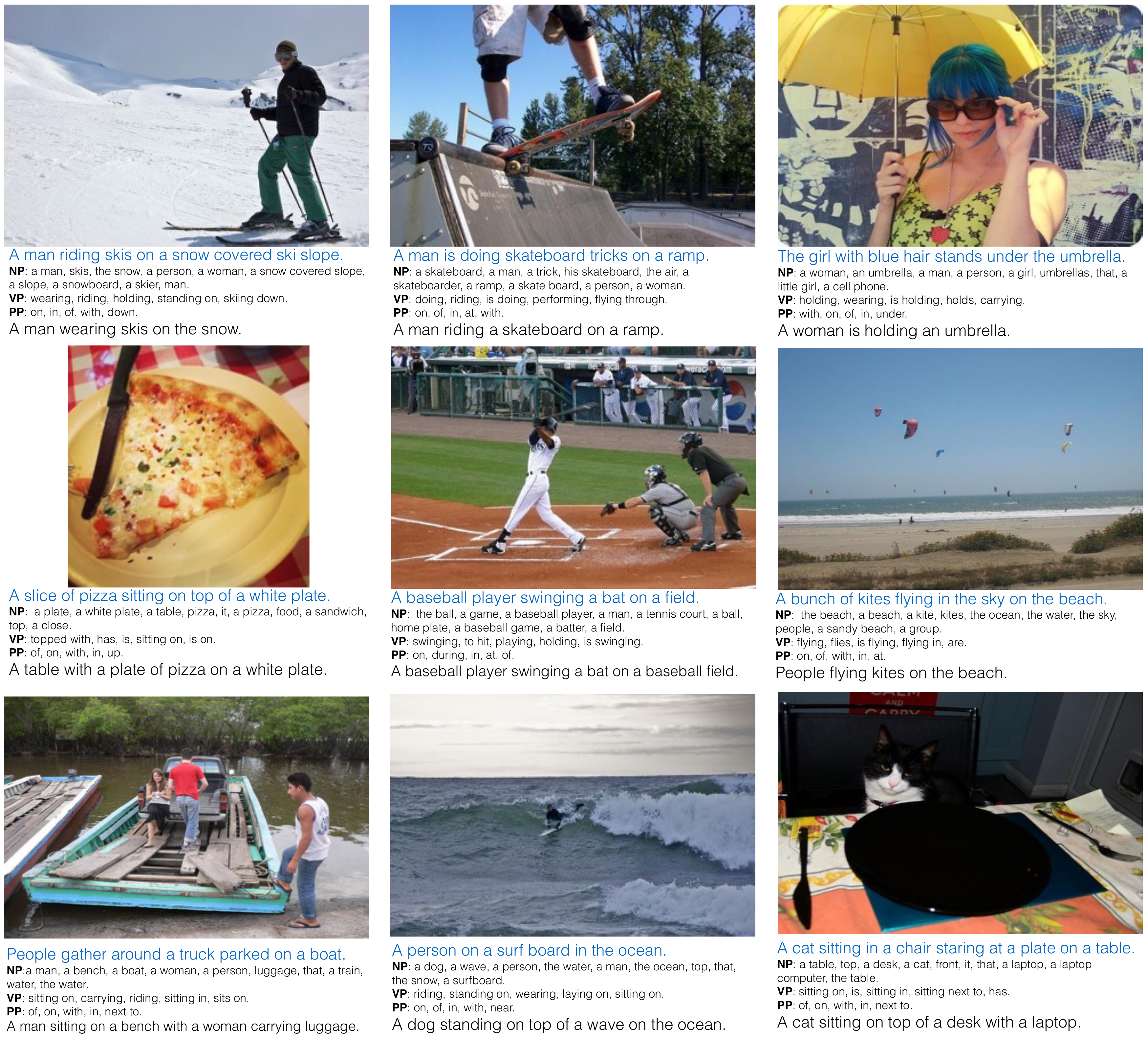}
  \end{center}
  \caption{Quantitative results for images on the COCO dataset. Ground-truth annotation (in blue), the NP, VP and PP predicted from the model and generated annotation (in black) are shown for each image. The last row are failure samples.}
  \label{fig:results}
\end{figure*}

We show examples of full automatic generated sentences in Figure~\ref{fig:results}.  The simple language model used is able to generate sentences that are in general syntactically correct. Our model produces sensible descriptions with variable complexity for different test samples. Due to the generative aspect of the model, it can occur that the sentence generated is very different from the ground-truth and still provides a descent description. The last row of Figure~\ref{fig:results} illustrates failure samples. We can see in these failure samples that our system has however outputted relevant phrases. 
There is still room for improvement for generating the final description. We deliberately choose a simple language model to show that competitive results can be achieved with a simple approach. A more complex language model could probably avoid these failure samples by considering a larger context. The probability for \emph{a dog} to stand on top of \emph{a wave} is obviously very low, but this kind of mistake cannot be detected with a simple trigram language model.

\subsection{Diversity of Image Descriptions}

In contrast to RNN-based models, our model is not trained to match a given image $i$ with its ground-truth descriptions $s$, i.e., to give $P(s|i)$.
Because our model outputs instead a set of phrases, this is not really surprising that only 1\% of our generated descriptions are in the training set for Flickr30k, and 9.7\% for COCO. 
While a RNN-based model is generative, it might easily overfit a small training data. 
\citet{VinyalsTBE14} report, for instance, that the generated sentence is present in the training set 80\% of the time.
Our model therefore offers a good alternative with the possibility of producing unseen descriptions with a combination of phrases from the training set. 

\subsection{Phrase Representation Fine-Tuning}

\begin{table}[h]
\vspace{-.8cm}
\resizebox{\columnwidth}{!}{
\begin{tabular}{@{}lcccc@{}}\hline\toprule
{\bf PHRASES} & & \multicolumn{2}{c}{\bf NEAREST NEIGHBORS} \\
\cmidrule{3-4}
 & \# & \sc before & \sc after \\\bottomrule
 \\
 \multirow{5}{1.95cm}{\sc a grey cat} & \small 1 & \sc\small a grey dog & \sc\small a gray cat\\
 & \small 2 & \sc\small a grey and black cat & \sc\small a grey and black cat \\
 & \small 3 & \sc\small a gray cat & \sc\small a brown cat \\
 & \small 4 & \sc\small a grey elephant  & \sc\small a grey and white cat \\
 & \small 10 & \sc\small a yellow cat & \sc\small  grey and white cat \\\bottomrule
 \\
 \multirow{5}{1.95cm}{\sc home plate} & \small 1 & \sc\small a home plate & \sc\small a home plate \\
 & \small 4 & \sc\small a plate & \sc\small home base\\
 & \small 6 &\sc\small another plate & \sc\small the pitch\\
 & \small 9 &\sc\small a red plate & \sc\small the batter \\
 & \small 10 &\sc\small a dinner plate & \sc\small a baseball pitch \\\bottomrule
 \\
 \multirow{5}{1.95cm}{\sc a half pipe} & \small 1 & \sc\small a pipe & \sc\small a pipe \\
 & \small 2 &\sc\small a half & \sc\small the ramp  \\
 & \small 5 &\sc\small a small clock & \sc\small a hand rail \\
 & \small 9 &\sc\small a large clock & \sc\small  a skate board ramp \\
 & \small 10 &\sc\small a small plate & \sc\small  an empty pool \\\bottomrule
\end{tabular}}
\caption{Examples of three noun phrases from the COCO dataset with five of their nearest neighbors before and after learning.}
\label{tab:neigh}
\vspace{-.8cm}
\end{table}

Before training the model, the matrix $U$ is initialized with phrase representations obtained from the whole English Wikipedia.
This corpus of unlabeled text is well structured and large enough to provide good word vector representations, which can then produce good phrase representations.
However, the content of Wikipedia is clearly different from the content of the image descriptions.
Some words used for describing images might be used in different contexts in Wikipedia, which can lead to out-of-domain representations for certain phrases. 
This becomes thus crucial to adapt these phrase representations by fine-tuning the matrix $U$ during the training\footnote{Experiments with a fixed $U$ phrase representations matrix significantly hurt the general performance. We observe about a 50\% decrease in both datasets with the BLEU metric. Since the number of trainable parameters is reduced, the capacity of $V$ should be increased to guarantee a fair comparison.}.
Some examples of noun phrases are reported in Table~\ref{tab:neigh} with their nearest neighbors before and after the training.
These confirm the importance of fine-tuning to incorporate visual features. 
In Wikipedia, \emph{cat} seems to occur in the same context than \emph{dog} or other animals. When looking at the nearest neighbors of a phrase such as \emph{a grey cat},  other \emph{grey} animals arise. After training on images, the word \emph{cat} becomes the important feature of that phrase. And we see that the nearest neighbors are now cats with different colours.
In some cases, averaging word vectors to represent phrases is not enough to capture the semantic meaning.
Fine-tuning is thus also important to better learn specific phrases. 
Images related to baseball games, for example, have enabled the phrase \emph{home plate} to be better defined. 
This is also true for the phrase \emph{a half pipe} with images about skateboarding.
This leads to interesting phrase representations, grounded in the visual world, which could be possibly used in natural language applications in future work.

\vspace{-.2cm}
\section{Conclusion}
\label{conclusion}
In this paper, we propose a simple model that is able to infer different phrases from image samples. From the phrases predicted, our model is able to automatically generate sentences using a statistical language model. We show that the problem of sentence generation can be effectively achieved without the use of complex recurrent networks. 
Our algorithm, despite being simpler than state-of-the-art models, achieves similar results on this task. Also, our model generate new sentences which are not generally present in training set.
Future research directions will go towards leveraging unsupervised data and more complex language models to improve sentence generation. 
Another interest is assessing the impact of visually grounded phrase representations into existing natural language processing systems.   

\section*{Acknowledgements}

This work was supported by the HASLER foundation through the grant ``Information and Communication Technology for a Better World 2020'' (SmartWorld).

\bibliography{icml2015}
\bibliographystyle{icml2015}

\end{document}